\documentclass{article}

\PassOptionsToPackage{numbers,square}{natbib}

\usepackage[preprint]{nips_2018}

\usepackage[utf8]{inputenc} 
\usepackage[T1]{fontenc}    
\usepackage{hyperref}       
\usepackage{url}            
\usepackage{booktabs}       
\usepackage{amsfonts}       
\usepackage{nicefrac}       
\usepackage{microtype}      
\usepackage{amsmath}
\usepackage{caption}
\usepackage{subcaption}
\usepackage{graphicx}
\usepackage[norelsize ]{algorithm2e}
\title{Kalman Filter Modifier for Neural Networks in Non-stationary Environments}

\author{
	Honglin Li \\
	CVSSP, University of Surrey\\
	\texttt{h.li@surrey.ac.uk} \\
	\And
	Frieder Ganz\\
	Adobe Systems Inc\\
	\texttt{ganz@adobe.com} \\
	\AND
	Shirin Enshaeifar\\
	CVSSP, University of Surrey \\
	\texttt{s.enshaeifar@surrey.ac.uk}\\
	\And
	Payam Barnaghi \\
	CVSSP, University of Surrey \\
	\texttt{p.barnaghi@surrey.ac.uk}\\
}

\begin{document}
	
	\maketitle
	
	\begin{abstract}
		Learning in a non-stationary environment is an inevitable problem when applying machine learning algorithm to real world environment. Learning new tasks without forgetting the previous knowledge is a challenge issue in machine learning. We propose a Kalman Filter based modifier to maintain the performance of Neural Network models under non-stationary environments. The result shows that our proposed model can preserve the key information and adapts better to the changes. The accuracy of proposed model decreases by 0.4\% in our experiments, while the accuracy of conventional model decreases by 90\% in the drifts environment.
		
	\end{abstract}
	
	\section{Introduction}
	In many real-world scenarios, the underlying process of data stream is non-stationary. The performance of neural networks may be decreased when the source distribution of the input changes. Even worse, these changes may lead to catastrophic forgetting problem. An important issue for neural networks is the ability to preserve previously learned information and to adapt faster to the changes in dynamic environment. We aim to solve the problem of forgetting previously learned information under data drifts. According to the Bayesian Decision Theory \cite{duda2001pattern} a classification problem can be defined as maximising the posterior probability of $P(y|X)$, where $y$ represents the classes of data ($X$). We can view the drifts into two types \cite{tsymbal2004problem,vzliobaite2010learning}: \textit{real drift} which refers to changes in $P(y|X)$ which means the data distribution remains the same, but the class of the data changes; \textit{virtual drift} which refers to the changes in $P(X)$ in which the class of the data remains the same, but the distribution of the data changes. Real drift and virtual drift need replacement learning and supplemental learning respectively\cite{elwell2011incremental}. In real world, we cannot choose the learning environment, a robust model is needed to adapt to these drifts. We propose a model with a Kalman Filter modifier to adjust the learning parameters of the neural network models. Our experiments show that our purposed solutions adapts better and faster compared to the conventional neural network models in drifts environment.
	\section{Kalman Filter Modifier}
\begin{figure}[h]
		\centering
		\begin{subfigure}[b]{0.3\textwidth}
			\includegraphics[width=\linewidth]{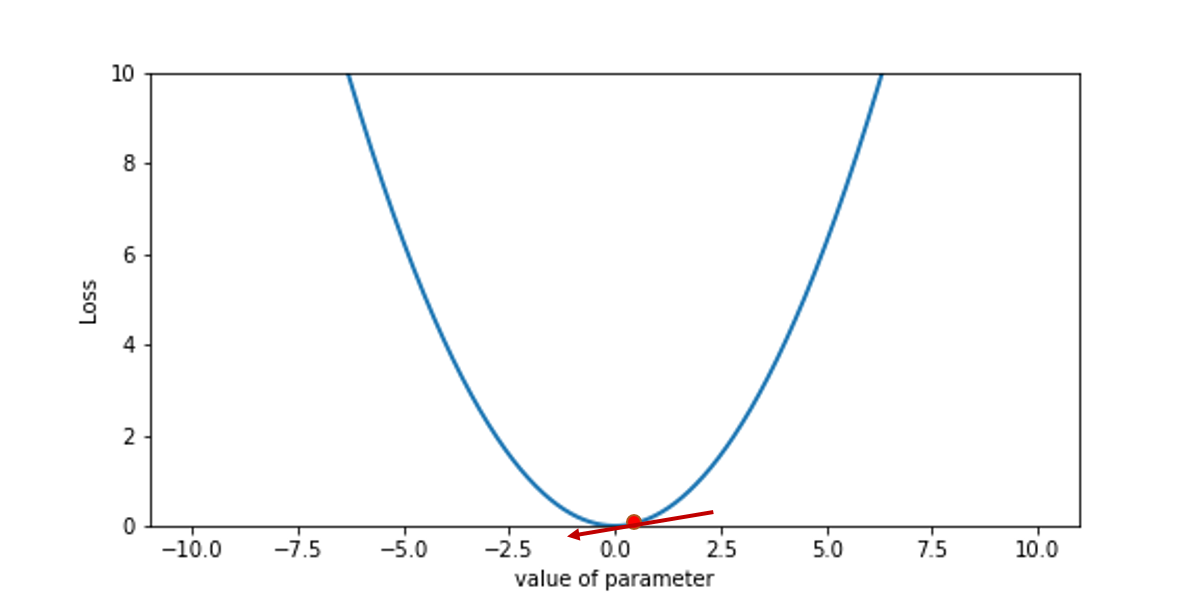}
			\caption{Optima Value of task A}
		\end{subfigure}
		\begin{subfigure}[b]{0.3\textwidth}
			\includegraphics[width=\linewidth]{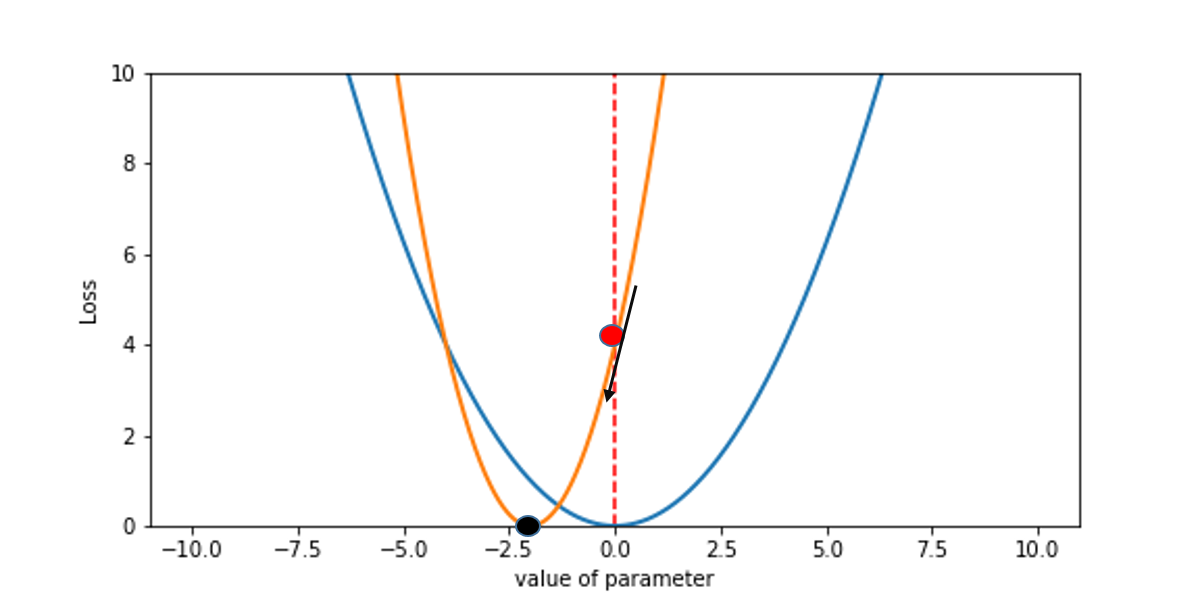}
			\caption{Model move from A to task B}
		\end{subfigure}
		\begin{subfigure}[b]{0.3\textwidth}
			\includegraphics[scale=0.5]{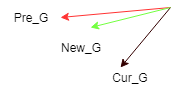}
			\caption{Estimate Gradient}
		\end{subfigure}
		\caption{(a) The parameter of the model will be around the red point when the model is converged. The red line indicates the gradient of the red point. (b) The training task changes from A to B. The red point is no longer an optima value. The black line shows the parameter has a tendency of moving to black point. (c) The red line(Pre\_G) is the gradient of the model on task A, the black one(Cur\_G) is the gradient of the model on task B. Our proposed method can estimate a new value(New\_G) based on these two gradients. }
		\label{fig:kal}
	\end{figure}

	The performance of an online learning model would decrease if the training data or its distribution changes. From the gradient point of view, the location of optima point will change. For example, if we take a single parameter as shown in Figure (\ref{fig:kal}), the parameters in the pre-trained model have a minimum loss. However, the loss will be relatively high when the distribution of the data changes. In this case, the parameters of the online learning model will change to the new optima value. However, this changes could be a significant problem in training a consistent model when the data periodically goes back to the previous state or distribution\cite{kirkpatrick2017overcoming}. 
	We address this problem by finding an optimal estimation between the data and the changes caused by the drifts. For this purpose we train a Kalman filter which acts as an optimal estimator\cite{rhudy2017kalman}. This method can infer parameters from uncertain an inaccurate observations.
	\begin{equation}
	\label{eq:states}
	\begin{split}
	m_k = Am_{k-1} + Bu_k + w_k, z_k = Hm_k + v_k
	\end{split}
	\end{equation}
	
	In our approach, we use a mini-batch method to train the model. In Equation (\ref{eq:states}) $m_k$ is the $k_{th}$ state of model, the initial state $m_0$ represents the pre-trained model. Each state means model is training on $k_{th}$ batch of the new data. $z_k$ refers to the output model. From the gradient descent algorithm view, in the Equation (\ref{eq:states}), $A = I$, $B$ is the learning rate, $u_k$ is the gradient of model on $D_k$ which refers to the $k_{th}$ batch of data, $H=I$. Because we omit the process noise, the $w_k,v_k$ are 0. This is how the neural networks perform with gradient descent algorithm in a linear state perspective.
	
	However, the model trained this way is not an optimal or accurate model. Based on the gradient algorithm, we can obtain the gradient of the pre-trained model which will indicate the measurement error of the current model. Using a Kalman Filter, we have the following:
	\begin{equation}
	\label{eq:predict}
	\begin{split}
	\hat{x}_{k|k-1} &= x_{k-1}  \\
	P_{k|k-1} &= P_{k-1} \\
	\end{split}
	\end{equation}
	We omit all the process noise, and assumes that the system is stable (the dynamic matrix is 0). In Equation (\ref{eq:predict}), the first formula is state predict process, the second one is state error predict process. Where $\hat{x_{k|k-1}}$ is the predicted model parameters at $k_{th}$ state, which we assumes it is stable with no additional information given, $P_k$ is the state error, which assumes it is stable as well. 
	\begin{equation}
	\label{eq:update}
	\begin{split}
	K_k &= P_{k|k-1} / (P_{k|k-1} + R) \\
	\hat{x_k} &= \hat{x}_{k|k-1} + K_k (z_k - \hat{x}_{k|k-1}) \\
	P_k &= (I - K_k) P_{k|k-1}
	\end{split}
	\end{equation}
	Equation (\ref{eq:update}) is the update process. Where $R$ is measurement noise (gradient of $m_k$ on $D_k$), $K_k$ is the Kalman Gain at $k_{th}$ state, $\hat{x_k}$ is the estimated parameters of the model. Algorithm (\ref{al:kal}) summarises the proposed steps.
	
	\begin{algorithm}[H]
		\SetAlgoLined
		\KwResult{Model $M_k$}
		initialisation\;
		$P_0$ = gradients($m_0$ on pre\_trained dataset)\\
		$m_0$ = Pre\_trained Model\\
		\While{train on new task}{
			Produce model $m_k$: $m_{k-1}$ Train on $D_k$\\
			Calculate $R$: $R$ = gradients($m_k$ on $D_k$)\\
			Kalman Gain: $K_k =  P_k / (P_{k-1} + R)$ \\
			Produce model $\hat{m_k}$: $\hat{m_k}$ = Kalman Filter([$m_{k-1}$,$P_{k-1}$],[$m_k$,$P_k$])\\
			Update $P_k = (I - K_k) P_k$
		}
		\caption{The Kalman Filter based Algorithm}
		\label{al:kal}
	\end{algorithm}
	
	\section{Experiment Results}
	
\begin{figure}[h]
		\centering
		\begin{subfigure}[b]{0.2\textwidth}
			\includegraphics[width=\linewidth]{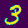}
			\caption{Original}
		\end{subfigure}\hspace{5mm}
		\begin{subfigure}[b]{0.2\textwidth}
			\includegraphics[width=\linewidth]{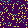}
			\caption{Permuted }
		\end{subfigure}\hspace{5mm}
		\begin{subfigure}[b]{0.2\textwidth}
			\includegraphics[width=\linewidth]{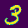}
			\caption{Label changed}
		\end{subfigure}
		\caption{(a) The pre-trained MNIST dataset. (b) Permuted MNIST dataset. (c) The training images are the same, but the labels are increased by one; in this example, the label of the image will be 4.}
		\label{fig:data}
	\end{figure}

\begin{figure}[h]
		\centering
		\begin{subfigure}[b]{0.4\textwidth}
			\includegraphics[scale=0.35]{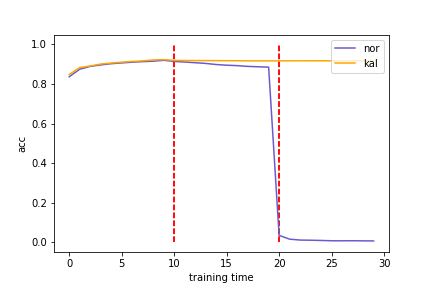}
			\caption{Progression of accuracy on validation data}
		\end{subfigure}
		\begin{subfigure}[b]{0.4\textwidth}
			\includegraphics[scale=0.35]{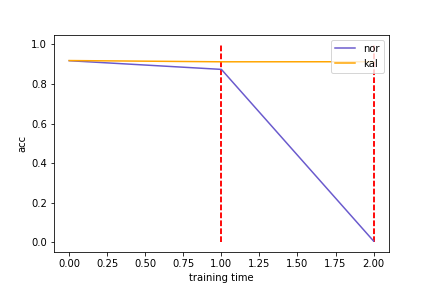}
			\caption{Progression of accuracy on test data }
		\end{subfigure}
		\caption{(a) The accuracy of the model using pre-trained validation dataset. (b) The accuracy of the model using pre-trained test dataset after being trained on new tasks. The oranges line are the accuracy of the proposed model, the blues line are the accuracy of conventional neural networks. The red dashed lines indicate the change point in the tasks. Each dashed line shows a drifts.}
		\label{fig:res}
	\end{figure}

	We train a fully-connected multi-layer neural network\cite{schmidhuber2015deep} on three datasets sequentially. Within each task, the model is trained at fixed epochs and the training data will be no longer available to the model. We constructed a set of classification tasks based on the MNIST dataset\cite{lecun2010mnist} (Figure \ref{fig:data}). The data in the first task is the original MNIST dataset. In the second task we permute all the pixels of the images. This will require a completely different solution. The final task is related to the real drift problem. For this purpose, we change all the labels by adding 1 to the value of the label (e.g. if the image is 3, the label will be changed to 4).  The results show that, no matter what the training dataset is, the Kalman filter will allow the online learning model to respond more efficiently to the changes and to maintain an overall better performance compared to a conventional model without any modifiers (see Figure \ref{fig:res}).
	

	\section{Conclusion}
	We present a novel online learning method that responds to data drifts by using a Kalman Filter modifier. This addresses the forgetting problem for neural networks in non-stationary environments. Our proposed method does not require any changes in the architecture of neural network. We use the Kalman filter to adjust the learning parameters in changing environments. The Kalman Filter modifier takes the weights as the measurement value and the gradient as the measurement error. We demonstrate our approach using both virtual and real drifts and show that the proposed model will remember the previously learned information to adjust the online learning parameters. The method is characterised by an intrinsic recursive algorithm; so it does not need to access the previously seen data. Our evaluation results show that our approach performs better in responding to changes and has lower learning error compared with a conventional model. Our future work focus on improving the Kalman Filter and compare it with some advanced catastrophic forgetting methodologies in non-stationary environment.
	
	\subsubsection*{Acknowledgments}
	This work is partially supported by the EU H2020 IoTCrawler project under contract number: 779852.
	\bibliography{ref}
	
\end{document}